%% file: main.tex
\documentclass[a4paper,conference]{IEEEtran}

\IEEEoverridecommandlockouts
\usepackage{cite}
\usepackage{amsmath,amssymb,amsfonts}
\usepackage{algorithmic}
\usepackage{graphicx}
\usepackage{textcomp}
\usepackage{xcolor}
\def\BibTeX{{\rm B\kern-.05em{\sc i\kern-.025em b}\kern-.08em
    T\kern-.1667em\lower.7ex\hbox{E}\kern-.125emX}}

\input{latexTools}
    
\begin{document}

\title{Self-Supervised Joint Encoding of Motion and Appearance for First Person Action Recognition
\thanks{Computational resources provided by hpc@polito (http://www.hpc.polito.it).}
}

\author{\IEEEauthorblockN{Mirco Planamente}
\IEEEauthorblockA{\textit{Dept. of Control and Computer Eng.} \\
\textit{Politecnico di Torino}\\
Torino, Italy \\
\textit{Italian Institute of Technology} \\
mirco.planamente@polito.it}
\and
\IEEEauthorblockN{Andrea Bottino}
\IEEEauthorblockA{\textit{Dept. of Control and Computer Eng.} \\
\textit{Politecnico di Torino}\\
Torino, Italy \\
andrea.bottino@polito.it}
\and
\IEEEauthorblockN{Barbara Caputo}
\IEEEauthorblockA{\textit{Dept. of Control and Computer Eng.} \\
\textit{Politecnico di Torino}\\
Torino, Italy \\
\textit{Italian Institute of Technology} \\
barbara.caputo@polito.it}
}

\maketitle

\input{abstract}

\input{introduction}

\input{soa}
\input{method}

\input{experiments}
\input{conclusions}

\bibliographystyle{IEEEtran}
\bibliography{biblio}

\end{document}

%% file: latexTools.tex
\usepackage{xcolor} 
\usepackage{soul} 
\usepackage{xspace}
\usepackage{amsmath}

\usepackage{hyphenat}
\newcommand{\mymethod}{SparNet\xspace}

\newcommand{\quotes}[1]{``#1''}

\usepackage{etoolbox}
\usepackage{tikz}

\newrobustcmd*{\mysquare}[1]{\tikz{\filldraw[draw=#1,fill=#1] (0,0)
rectangle (0.2cm,0.2cm);}}

\newrobustcmd*{\mycircle}[1]{\tikz{\filldraw[draw=#1,fill=#1] (0,0) circle [radius=0.1cm];}}

\newrobustcmd*{\mytriangle}[1]{\tikz{\filldraw[draw=#1,fill=#1] (0,0) --
(0.16cm,0) -- (0.08cm,0.16cm);}}

\newrobustcmd*{\mytriangledown}[1]{\tikz{\filldraw[draw=#1,fill=#1] (0,0.16) --
(0.16cm,0.16) -- (0.09cm,0cm);}}

\definecolor{darkgreen}{rgb}{0.0, 0.5, 0.0}
\definecolor{darkred}{rgb}{0.5, 0.0, 0.13}

%% file: abstract.tex
\begin{abstract}

Wearable cameras are becoming more and more popular in several applications, increasing the interest of the research community in developing approaches for recognizing actions from the first-person point of view. An open challenge in egocentric action recognition is that videos lack detailed information about the main actor's pose and thus tend to record only parts of the movement when focusing on manipulation tasks. Thus, the amount of information about the action itself is limited, making crucial the understanding of the manipulated objects and their context. Many previous works addressed this issue with two-stream architectures, where one stream is dedicated to modeling the appearance of objects involved in the action, and another to extracting motion features from optical flow. In this paper, we argue that learning features jointly from these two information channels is beneficial to capture the spatio-temporal correlations between the two better. To this end, we propose a single stream architecture able to do so, thanks to the addition of a  self-supervised block that uses a pretext motion prediction task to intertwine motion and appearance knowledge. Experiments on several publicly available databases show the power of our approach.

\end{abstract}

%% file: introduction.tex
\section{Introduction}
\label{sec:intro}
Recognizing human actions from videos is one of the most critical challenges in computer vision since its infancy. The capability to automatically (and reliably) recognize the action performed by an individual or a group of people would have a tremendous impact on a plethora of applications, ranging from security and surveillance to autonomous driving, automatic indexing and retrieval of media content, human-robot, and human-computer interaction, and many others. Historically, most of the work has been done on third-person action recognition, an area where good progress has been made and applications are already finding their way on the market. In the last years, the technological advances in the field of wearable devices led to a growing interest in first-person action recognition (FPAR) due to the possibility to capture activities following the user in mobility and without the need to place sensors in the environment.

\begin{figure*}[http]
\begin{center}
\includegraphics[clip,width=.8\textwidth]{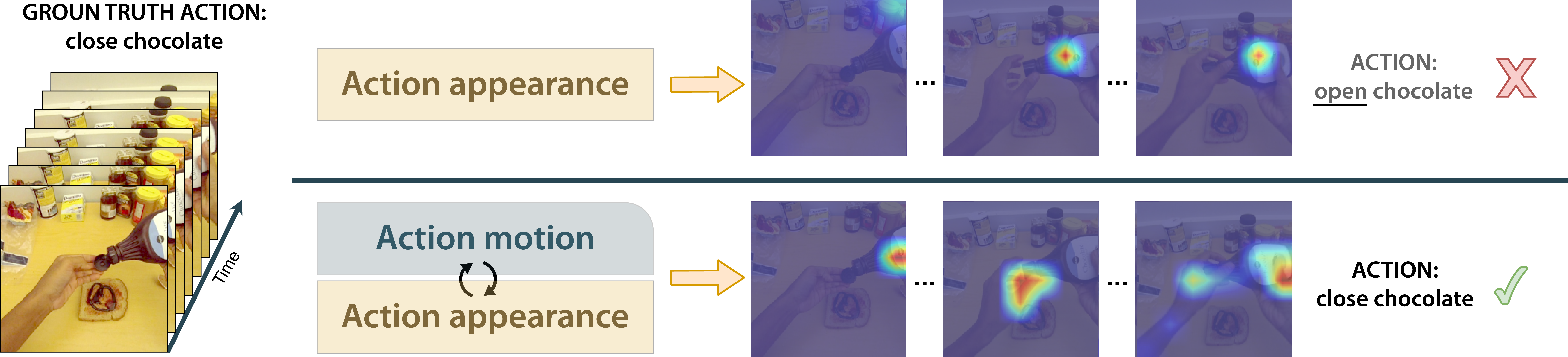}
\end{center}
   \caption{By leveraging a self-supervised motion prediction task during training, at test time, \mymethod can jointly exploit motion and appearance information using a single RGB stream. The result is a leaner architecture that better focuses on the relevant elements for action recognition in egocentric vision, as can be seen from the comparison between the Class Activation Maps when the auxiliary task is used (lower path) or not (upper path) }
\label{fig:teaser}
\end{figure*}

When moving from third-person to first-person action recognition, a first issue to face is how to deal with strong egomotions, as data are usually acquired by wearable cameras mounted on the actor body. A second and perhaps even more relevant challenge is the scarcity of available information about the pose of the main actor, as opposed to third-person videos. Most egocentric videos contain actions of the camera wearer interacting with objects \cite{ekECCV2018}, with only parts of the arm trajectory and the hand gestures visible in the captured data. Following this observation, it becomes crucial to extract from video frames as much information as possible on the objects being (or about to be) manipulated, their position and the motion data encoded in the video (since, for instance, the correct interpretation of the actions of \quotes{opening} and \quotes{closing} a bottle, Fig. \ref{fig:teaser}, depends merely on the hand motion direction).

A popular approach for addressing the last issue is to combine two  
pieces of information: the visual appearance of the object of interest, modeled by the spatial stream that processes RGB images, and the motion information, handled by the temporal stream that takes as input the optical flow extracted from adjacent frames (for a detailed discussion on previous work we refer to section \ref{sec:soa}). Effective 
approaches integrate the basic two-stream architecture with attention modules aimed at identifying the frames and the regions in the frames that are more informative for the task at end \cite{Sudhakaran_2019_CVPR,sudhakaran2018attention}.
Despite the good level of success obtained, these methods present two main disadvantages. First, appearance and motion features are learned separately, and the final predictions of the two streams are merged only at the end of the network using (usually) simple weighted sums \cite{Simonyan2014, sudhakaran2018attention, Sudhakaran_2019_CVPR}. However, this choice is sub-optimal since it does not model their correlated spatial-temporal relationships. Second, pushing the envelope in two-stream approaches results in a growth of the number of parameters of the overall architecture. As a consequence, optimization is often performed in multiple stages.

In this paper, we address these issues by moving beyond the two-stream paradigm and proposing an architecture that couples the modeling of motion and appearance information within a single RGB stream 
by leveraging one or more motion-prediction (MP) self-supervised tasks. These tasks \quotes{force} the backbone to learning an image embedding that focuses on object movements, a piece of information that is beneficial for the main task of FPAR. Thanks to the use of the auxiliary tasks, this information is directly encoded in the inner layers of the backbone, hence leading to an intertwined learning of appearance and motion features. The effectiveness of this idea is demonstrated not only by our results but also by those obtained including these MP pretext tasks in other recent models such as Ego-RNN \cite{sudhakaran2018attention} and LSTA \cite{Sudhakaran_2019_CVPR} (see Section \ref{sec:ablation}).
The resulting architecture is relatively simple, as it consists of a standard backbone (i.e., a ResNet-34 in our experiments), followed by a standard ConvLSTM, and the heads of the auxiliary tasks are composed by shallow architectures. Because of its simplicity, it can be trained end-to-end in a single stage, as opposed to several other two streams methods \cite{sudhakaran2018attention, Sudhakaran_2019_CVPR}.
Furthermore, it can use a smaller amount of frames than what done in previous works 
without any adverse effect on the performances.
 We call our architecture Self-supervised first Person Action Recognition network - \mymethod.

To summarize, the contributions of our paper are as follows:
(i) we introduce, for the first time, a set of motion prediction self-supervised tasks in the specific domain of egocentric action recognition; 
(ii) we address the problem of how to effectively leverage over a self-supervised branch to jointly encode spatial and motion information {by identifying the features that are most suited to solve this task}; 
and (iii) we showcase the effect of each component of \mymethod with a quantitative and qualitative ablation study.

In the rest of the paper, we first revise previous work in FPAR and self-supervised learning, and we discuss into detail how we position ourselves with previous approaches that relate to some extent with \mymethod (Section \ref{sec:soa}). Section \ref{sec:method} describes our proposed architecture, while experiments are presented in Section \ref{sec:experiments} and discussion on future work are reported in the conclusion.

%% file: soa.tex
\section{Related works}
\label{sec:soa}

\textbf{First Person Action Recognition.}
The literature on FPAR has long acknowledged that the motion of the hands, the appearance of the objects being used, and their interplay are the most critical characteristics to extract from raw data \cite{degeest2016,pirsiavash2012,ghirshick2015}.
This approach has moved from using handcrafted features to the deep learning wave \cite{ma2016,Li_2018_ECCV,sudhakaran2018attention}, providing researchers with powerful and effective models for encoding appearance. 
However, such approaches neglect the temporal relationships and the dynamics between frames. Thus, some works proposed to tackle this issue exploiting Convolutional Long Short-Term Memory (ConvLSTM) networks  \cite{Sudhakaran_2019_CVPR,sudhakaran2018attention,sudhakaran2017convolutional}. Other works  \cite{Simonyan2014,sudhakaran2018attention, kazakos2019epicfusion} addresses the two tasks of recognition from motion and recognition from appearance with different networks that are either combined with late fusion or at the decision level. 
Recent methods attempted to strengthen the temporal aspects of videos using attention mechanisms \cite{Zhang2018,sudhakaran2018attention,Sudhakaran_2019_CVPR,perezrua2020knowing,Lu2019TIP} generally cast within a two-stream framework, to find the most informative parts through single images (spatial attention) or video segments (temporal attention). Although they have shown a reasonable degree of success, the resulting architectures tend to increase their complexity and often need to be trained in two stages or more.  
Latest models focuses on 3D CNNs \cite{Baidu-EK2019,ActionBanks-CVPR2019,kapidis2019multitask}, which leverage convolutional kernels spanning both spatial and temporal dimensions to provide combined representational patterns that are beneficial for FPAR. As an alternative, some architectures couple 3D CNNs with two-stream approaches \cite{Munro_2020_CVPR,lu2019learning,Baidu-EK2019} to obtain a better characterization of both short and long-term spatio-temporal dependencies among frames.
Some works try to turn a 2D CNN into an efficient spatial-temporal features extractor introducing a shift module that allows the exchange of information with neighboring frames \cite{lin2019tsm,sudhakaran2020gate}.
Finally, we also report recent works that combine multiple modalities like RGB, flow, audio and hand data \cite{lu2019learning, kapidis2019multitask, kazakos2019epicfusion}.

\textbf{Motion-based Self-Supervised Tasks.}
Self-Supervised Learning \cite{SSLsurvey} has been recently introduced for learning visual features from unlabeled data. The choice of an auxiliary task that does not require human annotation of the data enables the network to encode a knowledge that proves beneficial as initialization when solving a classification problem on related data. Different authors have proposed several auxiliary tasks, either relying on original visual cues (e.g., translation, scaling, rotation \cite{gidaris2018unsupervised}, clustering 
\cite{caron2018deep}, inpainting \cite{pathakCVPR16context} and colorization), on information extracted from image patches \cite{Noroozi_2018_CVPR},
 or on regularities introduced by the temporal dimension in videos \cite{Wang_UnsupICCV2015}.
The use of motion cues in self-supervised learning has been first proposed in \cite{Pathak2017} to segment a static frame (in an unsupervised way) using motion data obtained from videos. Optical flow cues have been exploited in \cite{HoWRC15}, to learn the visual appearance of obstacles in a Micro Air Vehicle landing environment. 
Instead, \cite{Mahendran2018} \quotes{transfer} optical flow information to pixel embeddings so that their difference matches that between optical flow vectors of the same pixels. 
The architecture in \cite{wang2019selfsupervised} regresses the spatial and temporal statistics of motion and appearance to learn spatio-temporal features for video representation. The main difference with our approach is that \cite{wang2019selfsupervised} leverages 3D convolutions that, working on volumes, already include implicit motion information clues (beyond that instilled by the auxiliary task) that we can not benefit from in our method, as we use sparse and individual images as input. 
Finally, in \cite{munro2020multi} the optical flow information has been used in a multi-modal self-supervised adaptation framework to help align the representation of different domains.

\textbf{Joint Appearance and Motion Modelling.}
As we already stated, while two-stream approaches (leveraging either 2D or 3D CNNs and possibly attention mechanism) are a powerful method for obtaining features that tightly couple appearance and motion cues, these opportunities come at the cost of an increase in model complexity and number of parameters.

This observation led several researchers to introduce approaches aimed at integrating both information in simpler architectures.  Zhao and Snoek
\cite{Zhao_2019_CVPR} propose the use of a single stream by conditioning with optical-flow the video representations obtained from a single RGB stream. However, although being less complex than a full two-stream model, at test time, the method still requires the availability of optical flow data.
Other works \cite{lee2018motion, sun2018optical} attempt to jointly model appearance and motion into a single stream, introducing modules designed to exploit the temporal information better, but still leading to complex architectures.
Finally, MARS \cite{crasto2019mars} attempts to hallucinate the optical flow information within a 3D CNN to avoid to compute it at test time while preserving the performance of two-stream approaches. 

%% file: method.tex
\section{SparNet: Architecture Overview and Details}
\label{sec:method}

\begin{figure*}[t]
\begin{center}
\includegraphics[clip,width=1\linewidth]{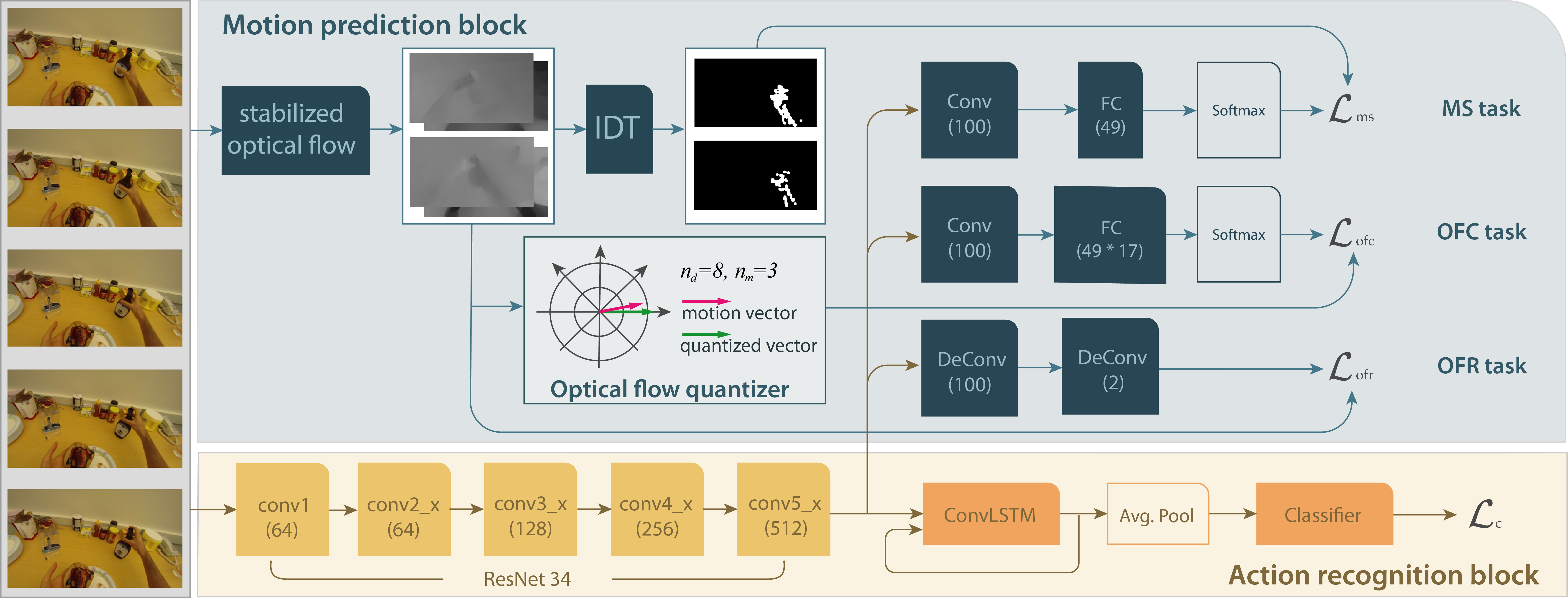}
\end{center}
  \caption{\mymethod architecture. The action recognition block computes image embeddings and solves the classification task. The motion prediction block injects  into the backbone motion information that results in a richer embedding, jointly encoding motion and appearance information. We compensate for egomotion effects with the use of stabilized dense optical flow and an IDT module \cite{Wang2013}. The OF quantizer block discretizes the input motion vectors into a finite number of classes according to the values of the $n_d$ and $n_m$ parameters (respectively, 8 and 3 in this example, for a total of 17 classes). }
\label{fig:architecture}
\end{figure*}

\textbf{\mymethod: Overview.}
In the basic version of the proposed architecture (Figure~\ref{fig:architecture}, action recognition block),  we first extract a small number $N$ of sparse representative RGB frames for each input video segment. The appearance embeddings of these frames, obtained by a standard CNN backbone, are fed to a ConvLSTM network whose output is first sent to an average pooling layer and then to a fully connected (FC) layer for classification. 
While processing a small number of frames helps reduce the computational burden of the model, the final features still lacks the motion information that is vital for the recognition process and that the two-stream approaches exploit by leveraging explicit optical flow data. 

To tackle this issue, we propose to regularize representation learning for FPAR by extending the basic architecture into a multi-task network. This network is required, at train time, to solve jointly two different problems: the action recognition task and a motion-prediction (MP) auxiliary task. We formalize the latter as a self-supervised problem that, given a single (and static) RGB frame as input, tries to answer one (or both) of the following questions: which parts of the image are going to move? And in which direction?

In our approach, we cast the identification of the moving parts, i.e., the motion segmentation task (MS), as a labelling problem aimed at minimizing the discrepancies between a \textit{motion map} (which labels pixels as either \textit{moving} or \textit{static}) and the object movements predicted by the network when observing a single static RGB frame. 
These unsupervised motion maps are obtained from the input video segment following the approach in \cite{Pathak2017} and exploiting the Improved Dense Trajectories (IDT, \cite{Wang2013}) to extract \quotes{stabilized} motion information. The main idea of IDT is first to compensate the strong camera motion and shake typical of egocentric videos by estimating the homography that relates adjacent frames. Then, the method labels as  \textit{moving} the keypoints that can be tracked reliably for at least eight frames and are not identified as camera motion.

The goal of the second MP sub-problem is to estimate the \quotes{stabilized} flow (i.e, the dense optical flow computed, as before, after compensating the camera motion between consecutive frames) from a static RGB input image. 
Since the optical flow is a continuous function, we first pose its estimation as a regression problem (OFR). As an alternative, we also turn it into a classification problem (OFC) by quantizing the per-pixel motion vectors in the following way. First, for each pixel, we extract the magnitude and direction of its motion vector. Then we discretize its angle into a set of $n_{d}$ directions uniformly distributed in the interval $[0, 2\pi]$. The magnitude is discretized by first clamping it to a maximal value $v_{clamp}$ and normalizing to one, and then dividing the interval $[0,1]$ into $n_{m}$ values (including the extremes), with the constraint  $n_{m} \geq 2$. Finally, each pair $(d_m, d_d)$ of discretized magnitude and direction values gets a unique label, except for all vectors with a magnitude close to zero (whose orientations tend to become meaningless) that are assigned to the same class. The total number of classes is, therefore, $n_{d} \cdot (n_{m} - 1) + 1$ and the self-supervised task has to estimate the correct flow labels.

Both approaches have their advantages and disadvantages. Regression is potentially more accurate, but it is also challenging to solve and, as noted in \cite{Walker2015ICCV}, it tends to smooth the results to the mean, ending in sub-optimal solutions. On the contrary, the classification problem has a more stable convergence, but it introduces quantization errors. Thus, a possible solution is combining the advantages of both methods. Similar benefits can be expected by coupling MS and optical flow estimation. MS identifies the points showing a stable and coherent motion in a temporal interval longer than that of two adjacent frames, thus, softening the effect of noise present in the dense optical flow. Conversely, optical flow estimation can provide the robust identification of the moving parts in an image with the information about the direction of their motion.

Despite the MP pretext task chosen (i.e., one of MS, OFR and OFC or any combination of them), its primary purpose is to help the appearance stream learn an embedding that encodes motion clues as well. We argue that, by processing an input with these characteristics, the ConvLSTM can extract a more meaningful global video representation (in terms of appearance and both short and long-term motion dependencies among frames) than the one observable from the vanilla appearance embeddings. 

\textbf{\mymethod: Details.}
Let $\mathcal{S}$ be a training set consisting of samples $S_i=\{H_i, y_i\}^{n}_{i=1}$, where $H_i$ is a set of $N$ timestamped images $\{(h^k_i,t^k_i)\}_{k=1}^{N}$ uniformly sampled from the video segment. Let also $x = f_M(H | \theta_f, \theta_c)$ be the embedding of sample $S$ computed by our model $M$, where parameters $\theta_f$ and $\theta_c$  define, respectively, the image embedding and the classification spaces. Finally, let $g(x)$ be a class probability estimator on the embedding $x$. 

The action recognition and the motion-prediction task share a common trunk that is completed by two task-specific “heads”. The first objective of the learning step consists in minimizing the categorical cross-entropy classification loss $\mathcal{L}_c$:
\begin{equation}
\mathcal{L}_{c}(x,y) = - \sum_{i=1}^{n} y_i \cdot log(g(x_i))
    \label{eq:lossClassification}
\end{equation}{}

Together with the objective mentioned above, we ask the network to solve an MP task, whose head can take different shapes according to the specific sub-problem chosen (or combination of sub-problems) and whose input is always the output of the backbone. 

The MS task is characterized by a shallow head composed by a single convolutional block, aimed at both adapting the features to the MS task and reducing their channel number. This head ends with a fully connected layer of size $s^2$  followed by a softmax, and it is trained with a loss $\mathcal{L}_{ms}$ based on the per-pixel cross entropy between the computed label image $l_{ms}$ and the ground truth $m$ (which is first downsampled to a size $s \times s$ and then vectorized). The estimated motion map $l_{ms}$ is obtained as a function of both image embedding $z$, which depends only on $\theta_f$, and MS head parameters ($\theta_{ms}$). Thus, the $\mathcal{L}_{ms}$ loss can be defined as:
\begin{equation}
 \mathcal{L}_{ms}(z,m) = - \sum_{i=1}^{n} \sum_{k=1}^{N}  \sum_{j=1}^{s^2}  m^k_i(j) \cdot log (l^k_{ms,i}(j))
    \label{eq:lossMotionSegmentation}
\end{equation}{}
where $m$ is the ground truth. 

The OFR task aims at regressing the two separate horizontal and vertical components of the optical flow. Its head is composed by a stack of two deconvolution layers (and ReLU activation functions) that learn a nonlinear upsampling to a final size of $r \times r \times 2$. The head is trained by minimizing the Mean Squared Error (MSE) between the predicted optical flow (which is a function of $\theta_f$ and OFR head parameters, $\theta_{ofr}$) and the ground truth (which is downsamples to a size $r \times r \times 2$). The $\mathcal{L}_{ofr}$ task loss can be defined as:
\begin{equation}
 \mathcal{L}_{ofr}(z,of) = \sum_{i=1}^{n} \sum_{k=1}^{N}  \frac{1}{r^2}\sum_{j=1}^{r^2}  ||of^k_i(j) - y^k_i(j)||^2_2
    \label{eq:lossOFRegression}
\end{equation}{}
where $of$ is the ground truth and $y$ is the estimated optical flow.

Finally, the structure of the OFC head is identical to that of the MS task, with the only exception that its ground truth is obtained by first downscaling the optical flow to a size $s \times s$ and then quantizing it. The OFC loss is then defined as:

\begin{equation}
 \mathcal{L}_{ofc}(z,of) = - \sum_{i=1}^{n} \sum_{k=1}^{N}  \sum_{j=1}^{s^2}  Q(of^k_i)(j) \cdot log (l^k_{ofc,i}(j))
    \label{eq:lossOFClassification}
\end{equation}{}
where $l_{ofc}$ is the computed label image (depending from $\theta_f$ and the OFC head parameters $\theta_{ofc}$) and $Q$ is the one-hot vector containing the result of the quantization of the $of$ ground truth.

As shown in Figure \ref{fig:architecture}, the optimal model of \mymethod is obtained by jointly solving the two separate optimization problems of minimizing $\mathcal{L}_c$ and the loss of the chosen MP problem, each equipped with an $L^2$ weight regularization term. With this architecture, the weights of the MP head are affected only by the backpropagation of its loss error, and the classification parameters ($\theta_c$) are affected only by the $\mathcal{L}_c$ error. Then both losses contribute to updating the weights of the backbone ($\theta_f$) through a weighted combination of their gradients (whose weights are a hyperparameter of the method). When multiple auxiliary tasks are used, each of them is optimized independently and (again) their weighted gradients are combined with that of the action classification head to optimize the backbone.

\textbf{\mymethod: Implementation.}
While \mymethod network can leverage over many possible  
backbones, we choose for our experiments a ResNet-34 model pre-trained on ImageNet. 
The motion-prediction heads receive in input the features extracted from the \texttt{conv5\_x} block of the ResNet (whose size is $7 \times 7 \times 512$). The MS and OFC heads reduce the feature channels to 100 and the size $s^2$ of their resulting ground truth is, therefore, $49$. Both OFR deconvolutional blocks have a $3 \times 3$ kernel and a stride of 2. The first reduces the input feature channels to 100 and the second to two (i.e., the estimated horizontal and vertical optical flow displacements). The final value of $r$ is 35. As for the ground truths, we compute the dense optical flow with the Gunnar-Farneback method \cite{FarnebackGunnar}. 

%% file: experiments.tex
\section{Experiments}
\label{sec:experiments}

In this Section, we first introduce the datasets used in our experiments, along with some implementation details. 
Then,  we conduct an ablation to show the effectiveness of the proposed self-supervised MP tasks and our single-stream approach, along with an analysis on their effect on different models. Finally, we discuss the results, which show the strength of \mymethod in the analyzed benchmarks.

\subsection{Datasets}
\label{sec:datasets}

We evaluated the proposed approach on four standard first-person action recognition datasets. 
GTEA-61 \cite{Fathi2011} 
includes 7 daily activities performed by 4 different subjects. Extended GTEA Gaze+ (EGTEA+, \cite{Li_2018_ECCV}) subsumes GTEA-61 and contains 
about 10,000 video segments from 86 different sessions annotated with 106 fine-grained actions. FPHA \cite{FirstPersonAction_CVPR2018} includes videos belonging to 45 different activity categories performed by 6 actors.
Finally, EK (the largest of all these datasets \cite{ekECCV2018}) contains about 40,000 video segments 
depicting hundreds of daily actions performed by 32 volunteers in their kitchen. 

\subsection{Implementation details}
\label{sec:trainingImplementationDetails}
\mymethod is trained end-to-end on a single stage. 
The ConvLSTM cell has 512 hidden units for temporal encoding and is initialized as in \cite{sudhakaran2018attention}. During training, we use different learning rates for the various architectural blocks (backbone, MS head, ConvLSTM, and final classification layer). The training epochs are 400 for GTEA-61, 70 for EGTEA+, 100 for FPHA and EK and we use ADAM as the optimization algorithm. Batch size is 4 for GTEA-61 and 8 for the remaining datasets. 

Each input video segment is decomposed into $N=7$ frames for GTEA-61 and FPHA and $N=11$ for EGTEA+ and EK, uniformly sampled in time. We fixed $v_{clamp}=15$ and $n_m=4$ with a varying number of angular subdivisions ($n_d \in \{8,16,20\}$). 
We resized input images at the height of 256 pixels, maintaining the same height ratio to update the width. The actual training input is a random crop of size $224 \times 224$ pixels. Ground truth for MP tasks was computed by first scaling all videos at a fixed height of 540 pixels. 
During training, we use the data augmentation techniques proposed in \cite{WangTPAMI2019}. At test time, we feed the network with the central crop of the frames.  
In order to foster result reproducibility, we made three runs for each experiment, each with the same constant seed across different datasets and parameters. Therefore, unless stated otherwise, \mymethod results are presented as the average accuracy over these three runs.

\input{Table/allResults}

\input{ablation}

\subsection{Experiments on GTEA-61, EGTEA+ and FPHA} 
\label{sec:egteaResults}

Experiments on GTEA-61 and EGTEA+ followed the protocols defined in \cite{sudhakaran2018attention,Sudhakaran_2019_CVPR}, which require to report the final average accuracy over different and fixed non-overlapping training and test sets. As for FPHA, we followed the 1:1 protocol that defines fixed training and test sets. 
We compare \mymethod with several state-of-the-art methods based on different approaches, i.e., one (or a combination) of two-stream \cite{sudhakaran2018attention, Sudhakaran_2019_CVPR,ma2016going,WangECCV2016,lu2019learning} or multi-stream \cite{furnari2020rolling} models, attention modules \cite{sudhakaran2018attention, Sudhakaran_2019_CVPR,Zhang2018,lu2019learning}, 3D CNN \cite{kapidis2019multitask,lu2019learning}, multi-task learning \cite{tekin2019h+,kapidis2019multitask} and exploiting hand posture data \cite{nguyen2019neural,tekin2019h+,kapidis2019multitask,GramMatrix}.

From Table~\ref{table:allResults}, it can be seen that \mymethod reaches the state of the art in all the benchmarks and experimental protocols. We think that these results are a clear indication that the motion clues induced in the (single) appearance stream by the MP tasks were indeed capable of improving the discriminative capabilities of the final embeddings, to an extent higher than that provided by using explicit optical flow information or 3D CNN, and without the need to include specific attention modules or using additional (supervised) information. As for the MP tasks considered, these results confirm the ablation ones, i.e., the optimality of the combined MP tasks and the lower contribution of OFR.

\subsection{Experiments on EK}
\label{sec:ekExperiments}

Since by the time of preparing this submission, the EK challenge closed and test labels were not yet available, we followed the experimental protocol proposed in \cite{kapidis2019multitask}, which mirrors the \textit{unseen} kitchen split of EK challenge defining a custom training (participants 1-29) and validation set (participants 30-31). 
We underline that the extremely challenging nature of EK pushed researchers to experiment with very complex architectures involving various combinations of 3D convolutions and prior supervised knowledge \cite{Baidu-EK2019,ActionBanks-CVPR2019,kapidis2019multitask}, multi-stream approaches \cite{sudhakaran2019fbkhupba,kazakos2019epicfusion,alej2019seeing} and ensemble methods \cite{Baidu-EK2019,sudhakaran2019fbkhupba,kazakos2019epicfusion} that make it difficult to appreciate the individual contribution of their components.

On the contrary, our choice was that of making the minimal architecture adjustment necessary to deal with the specific EK requirements. To this end, we added two separate FC layers at the end of the ConvLSTM for \textit{verb} and \textit{noun} and combined their outputs to define the action class. We used as loss the average categorical cross-entropy of all tasks. The rationale of this choice was to truly show how far a single, simple, yet effective approach could achieve in this setting.

Table \ref{table:EKResults} compare our top-1 results using as MP task the combination of MS and OFC (the most effective combination according to the results in Table \ref{table:allResults}) with that of \cite{kapidis2019multitask}, the only other method available running the same experimental settings. As can be seen, the two approaches are comparable, with \mymethod obtaining better accuracy on \textit{verbs} and lower ones on \textit{noun} and \textit{action}. These results partially contradict those obtained with EGTA+ where \mymethod was the best performer among the two in all splits. Possible explanations of these performance differences are the following. First, EK contains untrimmed video segments, which made our uniform frame subsampling scheme a sub-optimal choice. Then, even though EK videos are sensibly longer than that of other datasets, we had to pick only 11 representative frames (a number probably not optimal) due to the lack of sufficient computational and memory resources. Finally, we underline that \cite{kapidis2019multitask} leverage as backbone a Multi-Fiber Network, a 3D CNN pretrained on Kinetics and, thus, better suited for video processing than our ResNet-34.
That said, it is interesting to note the differences in \textit{verb} and \textit{noun} between the two methods. We think that the better accuracy of \mymethod on \textit{verb} is because the MP task helps our method focus on movements, while the hand position regression task of \cite{kapidis2019multitask} help their network better focus on the hand regions and, thus, implicitly on the objects the hands interact with.  Then, our \quotes{naif} way of combining \textit{verb} and \textit{noun} explains the final lower accuracy on \textit{action}. 

\begin{table}[tbp]
\caption{Top-1 accuracy on the EK validation set defined in \cite{kapidis2019multitask}  
}
\label{table:EKResults}
\centering
\begin{tabular}{p{3cm}ccc}
\hline
\hline
Method                      & Verb & Noun  & Action \\ 
\hline
3DConv MTL \cite{kapidis2019multitask}        
&  49.31 &  \textbf{27.60}   & \textbf{19.29}   \\
\mymethod-MS+OFC        
& \textbf{52.32} & 26.01 & 16.95          \\  

\hline 
\hline
\end{tabular}

\end{table}

%% file: Table/allResults.tex

\begin{table*}[ht]
\caption{Comparison with the state of the art on GTEA-61, EGETA+ and FPHA datasets. Best results are highlighted in bold}
\label{table:allResults}
\centering
\resizebox{.75\textwidth}{!}{%
\begin{tabular}{lcllcllc}
\cline{1-2} \cline{4-5} \cline{7-8} 
\noalign{\vskip\doublerulesep\vskip-\arrayrulewidth}
\cline{1-2} \cline{4-5} \cline{7-8} 
\noalign{\vskip\doublerulesep\vskip-\arrayrulewidth}
\multicolumn{2}{c}{GTEA-61} &  & \multicolumn{2}{c}{EGTEA+} &  & \multicolumn{2}{c}{FPHA} \\ 
\cline{1-2} \cline{4-5} \cline{7-8} 
\noalign{\vskip\doublerulesep\vskip-\arrayrulewidth}

EleAttG \cite{Zhang2018}       & 66.77 &  & RULSTM \cite{furnari2020rolling}  & 60.20 &  & H+O \cite{tekin2019h+}  & 82.43 \\
TSN \cite{WangECCV2016}             & 69.93 &  & Ego-RNN \cite{sudhakaran2018attention} & 60.76 &  & Gram Matrix \cite{GramMatrix}  & 85.39 \\
Ma et al. \cite{ma2016going}     & 73.02 &  & LSTA \cite{Sudhakaran_2019_CVPR}   & 61.86 &  & ST-TS-HGR-NET \cite{nguyen2019neural}      & 93.22 \\
Ego-RNN \cite{sudhakaran2018attention}           & \multicolumn{1}{l}{79.00}       &  & 3DConv MTL \cite{kapidis2019multitask}   & 65.70    &  &                  &                                \\
LSTA \cite{Sudhakaran_2019_CVPR}          & \multicolumn{1}{l}{80.01} &  & Two-stream I3D + STAM  \cite{lu2019learning} & 65.97 &  &                 &       \\ 

\noalign{\vskip\doublerulesep\vskip-\arrayrulewidth}
\cline{1-2} \cline{4-5} \cline{7-8} 
\noalign{\vskip\doublerulesep\vskip-\arrayrulewidth}

Baseline  & 80.18 &  & Baseline & 63.96 &  & Baseline & 94.32 \\
SparNet-MS  & 80.51 &  & SparNet-MS & 66.15 &  & SparNet-MS & 96.41 \\
SparNet-OFR & \multicolumn{1}{l}{80.14} &  & SparNet-OFR & 64.22 &  & SparNet-OFR & \multicolumn{1}{l}{95.07}     \\
SparNet-OFC    & 81.17 &  & SparNet-OFC & 67.36 &  & SparNet-OFC     & 96.41 \\
SparNet-OFR+OFC  & 80.51 &  & SparNet-OFR+OFC & \textbf{67.52}  &  & SparNet-OFR+OFC  & 96.35 \\
SparNet-MS+OFC & \textbf{81.39} &  & SparNet-MS+OFC & 67.44 &  & SparNet-MS+OFC  & \textbf{96.70} \\ 

\noalign{\vskip\doublerulesep\vskip-\arrayrulewidth}
\cline{1-2} \cline{4-5} \cline{7-8} 
\noalign{\vskip\doublerulesep\vskip-\arrayrulewidth}
\cline{1-2} \cline{4-5} \cline{7-8} 

\end{tabular}%
}
\end{table*}

%% file: ablation.tex
\subsection{Ablation Study}
\label{sec:ablation}

In this section, we comprehensively evaluate \mymethod on the first split of the EGTEA+ dataset using as baseline
the action recognition block in Figure~\ref{fig:architecture}. Specifically, we study the following aspects.

\begin{table}[htbp]
\caption{Number of representative frames and input features}
\label{table:ablationFrameFeatures}
\centering
\begin{tabular}{lccc}
\hline
\hline
Method                      & Accuracy (\%) \\ 
\hline
\mymethod-MS (7 frames)         & 67.05       \\
\mymethod-MS (11 frames)        & \textbf{68.43}           \\ 
\mymethod-MS (16 frames)        & 67.48           \\ 
\hline
\mymethod-MS @ \texttt{conv4\_x}          & 66.15                 \\ 
\mymethod-MS @ \texttt{conv5\_x}          & \textbf{68.43}       \\
\mymethod-MS @ \texttt{Output ConvLSTM}          & 67.68                 \\ 

\hline 
\hline
\end{tabular}

\end{table}

\textbf{Sparse sampling.}
We start by analyzing the effect of the number $N$ of input frames used for action recognition (in both train and test). 
Using as reference the MS task and varying $N$ in the interval $[5,25]$, we did not observe significant differences for values between 9 and 11, while the error started (slightly) increasing for smaller and higher amounts (see Table~\ref{table:ablationFrameFeatures}, where we only report a selection of significant values). Since these results are consistent with those obtained with other MP tasks (which we do not show for the sake of brevity), we conjecture that, as also observed in \cite{WangECCV2016}, a dense temporal sampling results in highly redundant information that is unnecessary for capturing the temporal dynamic of the video. Conversely, a too small number of frames can cause the loss of relevant cues for the current action. That said, in our main experiments, we heuristically adapted $N$ to the average segment length of the analyzed dataset.

\textbf{MP taks: input features.}
The effectiveness of our auxiliary task depends primarily on the features it receives in input. A \quotes{natural} option for a ResNet backbone is using the output features of its residual blocks. Another possible choice is exploiting the spatio-temporal representations obtained from the ConvLSTM. As can be seen from Table~\ref{table:ablationFrameFeatures}, \texttt{conv5\_x} features largely improve those from lower layers (see \texttt{conv4\_x} accuracy, results from other blocks were significantly lower and omitted). We think that this is a clear indication that MP tasks benefit from leveraging high-level and more structured information for their analysis. As for the lower accuracy obtained by the ConvLSTM features, we hypothesize that the spatio-temporal processing capability of ConvLSTM makes the solution of the MP task easier. This reduces the need for the backbone to incorporate further motion information in its embeddings with negative effects on the main FPAR task.

\begin{table}[htbp]
\caption{Ablation results on EGTEA+ (split 1, $N=11$) 
}
\label{table:ablation}
\centering
\footnotesize
\begin{tabular}{lcccc}
\hline
\hline
Method                      & Acc (\%) & Param (M) & GFLOPS\\ 
\hline
baseline          & 65.46   &  24.34 &  41.52 \\ 
\hline
\mymethod-MS                    & 68.43 &  24.87  & 41.55 \\ 
\mymethod-OFR                   & 65.73         &  24.80   & 43.01  \\ 
\mymethod-OFC ($n_d=8$)   & 69.32         &  30.39    & 41.61  \\ 
\mymethod-OFC ($n_d=16$)   & 69.49         &  36.16    & 41.68  \\ 
\mymethod-OFC ($n_d=20$)   & 68.96         &  39.04    & 42.21  \\ 
\mymethod-OFR+OFC ($n_d=16$)   & 69.57         & 36.62    & 43.17  \\ 
\mymethod-MS+OFC ($n_d=16$)   & \textbf{69.80}         &  36.70    & 41.71  \\ 
\hline
Ego-RNN RGB \cite{sudhakaran2018attention}         & -     & 24.34 &  94.36       \\ 
Ego-RNN  \cite{sudhakaran2018attention}            & -    & 45.71 &  98.31       \\ 
LSTA RGB  \cite{Sudhakaran_2019_CVPR}          & 57.96    & 41.22 &  114.92       \\ 
LSTA    \cite{Sudhakaran_2019_CVPR}            & 61.86    & 62.59 &  118.86       \\ 
Two-stream I3D + STAM  \cite{lu2019learning}        & 68.60    & - &  -       \\ 
3DConv MTL \cite{kapidis2019multitask}                & 68.99    & - &  -       \\ 
\hline 
\hline
\end{tabular}
\end{table}

\textbf{Impact of the MP tasks.}
To analyze the (individual and mutual) contribution of the different MP tasks, in Table~\ref{table:ablation} we report the average accuracy obtained by different variants of \mymethod along with the total number of parameters and GFLOPS of the resulting architecture. The table reports as well state of the art results on the same split.

This ablation clearly highlight the contribution of the MP tasks. Both individual and combined tasks improve the baseline, although to different extents. OFR confirms to be the most difficult task to solve and, thus, the one providing the lower contribution to the final FPAR task. Conversely, the choice of translating the OF estimation into a classification problem showed to be effective. It is also interesting to note 
that the mutual contribution of individual tasks helps making more robust the MP problem, which in turns provide a better integration of appearance and motion information in the backbone.



Concerning the computational burden of the MP tasks, we can say that their effect is in general minimal, exception made for OFC that requires a larger number of parameters for the classification (but still a limited increase in terms of GFLOPS). We recall that the baseline numbers are those required at test time (when the MP tasks are disabled). It can be seen that the relative increase in term of parameters (GFLOPS) is $2.2\%$ ($0.1\%$) and $1.9\%$ ($3.6\%$) for, respectively, MS and OFR and up to $60.41\%$ ($0.5\%$) for OFC (when $n_d=20$). These numbers can be compared by those expressed by Ego-RNN and LSTA, whose GFLOPS are substantially higher (an increase between $127.3\%$ to $186.3\%$) in both their single and two-stream versions and in both train and test time.

\textbf{MP tasks and other models.} 
One possible question is if the proposed MP  tasks can be beneficial to other models too. To this end, we performed a detailed analysis of their effects on Ego-RNN RGB \cite{sudhakaran2018attention} and LSTA-RGB \cite{Sudhakaran_2019_CVPR}. Both methods converge to the same baseline of \mymethod when the CAM is deactivated (in Ego-RNN RGB) or a vanilla LSTM cell is used instead of the proposed LSTA cell (in LSTA-RGB). For these experiments, we modified both architectures adding various MP tasks, feeding them with the \texttt{conv5\_x} features and 25 frames in input, as in their original papers. 

We present results obtained on the split 2 of GTEA-61. For the sake of brevity we report the results obtained with MS and OFC. 
For a fair comparison, single-stream results are those obtained in our experiments, which, despite our efforts, could not replicate those presented in \cite{sudhakaran2018attention} and \cite{Sudhakaran_2019_CVPR}.  We also underline that, since both methods retrain merely the last residual block of the backbone and not the whole ResNet as in our case, the MP effect is not back-propagated to the lower backbone layers, preventing them from supporting the higher ones in learning new features that are more focused on the actual FPAR task. 
Nonetheless, we think the numbers in Table~\ref{table:comparisonInAblation} highlight the effect of MP on these models and
showing that the effectiveness of our approach is not limited to \mymethod. 

\begin{table}[tbp]
\caption{Contribution  of  MS  task  on Ego-RNN  and  LSTA  
}
\label{table:comparisonInAblation}
\centering
\begin{tabular}{lccc}
\hline
\hline
Method                      & Single-stream (SS) & SS + MS & SS + OFC \\ 
\hline
Ego-RNN \cite{sudhakaran2018attention}%
&  63.79 &  68.97   & 68.10   \\
LSTA \cite{Sudhakaran_2019_CVPR}        
& 65.80 & 66.96 & 67.24          \\  

\hline 
\hline
\end{tabular}

\end{table}

%% file: conclusions.tex
\section{Conclusions}
\label{sec:conclusions}
In this paper, we presented \mymethod, a single stream architecture for FPAR. Its main feature is the ability to jointly learn appearance and motion features thanks to the use of a set of self-supervised pretext tasks aimed at estimating the motion information associated with a single static input image. This leads to a light architecture, trainable in a single stage, capable of working on sparse sampling of the input video segment (thus reducing test processing time) and achieving optimal results on several publicly available datasets.

Despite the promising results obtained, there is still room for further improvements. As future works, we are planning to investigate the contribution of other self-supervised pretext tasks and the possible integration of multiple auxiliary tasks, capable of further strengthening the discriminative capabilities of the final embeddings. Another option we are interested in is verifying the possibility of using modality hallucination approaches as an alternative to the MP tasks for instilling a \quotes{flavour} of motion into the (single stream) appearance features.